\documentclass[journal]{IEEEtran}
\ifCLASSINFOpdf
\else
\fi
\usepackage[cmex10]{amsmath}
\usepackage{mathtools}
\usepackage{amsfonts} 
\usepackage{siunitx}
\usepackage{algorithm,algorithmic}
\usepackage{booktabs}
\usepackage{threeparttable}
\usepackage{mathrsfs,amsmath}
\usepackage{multirow}
\usepackage{makecell}
\usepackage{placeins}
\usepackage{fixltx2e}
\usepackage{color}
\usepackage{isomath}
\usepackage{caption}
\usepackage{color}

\usepackage[numbers,sort&compress]{natbib}

\newcommand{\R}{\mathbb{R}}
\newcommand{\N}{\mathcal{N}}
\newcommand{\Sp}{\mathcal{S}}

\newcommand{\T}{\mathcal{T}}
\newcommand{\Lno}{\mathcal{L}}

\newcommand{\D}{\mathbfit{D}}

\newcommand\norm[1]{\left\lVert#1\right\rVert}
\DeclareSIUnit\Molar{\textsc{m}} %captial M for molar in SI{}{}
\DeclareSIUnit{\pH}{pH}
\DeclareSIUnit{\pixel}{px}
\renewcommand{\vec}[1]{\text{\boldmath$#1$}} 

\DeclareMathOperator*{\argmax}{\arg\!\max}
\begin{document}
%
% paper title
% can use linebreaks \\ within to get better formatting as desired
\title{Deep Learning and Conditional Random Fields-based Depth Estimation and Topographical Reconstruction from Conventional Endoscopy}
%
%
% author names and IEEE memberships
% note positions of commas and nonbreaking spaces ( ~ ) LaTeX will not break
% a structure at a ~ so this keeps an author's name from being broken across
% two lines.
% use \thanks{} to gain access to the first footnote area
% a separate \thanks must be used for each paragraph as LaTeX2e's \thanks
% was not built to handle multiple paragraphs
%

\author{Faisal Mahmood and Nicholas J. Durr
\thanks{F.M. and N.J.D. are with the Department of Biomedical Engineering, Johns Hopkins University (JHU), 3400 N Charles St., Baltimore, MD 21218, USA. \{faisalm, ndurr\}@jhu.edu}
%\thanks{Correspondence should be addressed to ulf.skoglund@oist.jp or faisalm@jhu.edu}% <-this % stops a space
%\thanks{This work was supported by Japanese Government OIST Subsidy for Operations (Skoglund U.) Grant No. 5020S7010020. F.M. and M.T were additionally supported by the OIST PhD Fellowship.}
%\thanks{Initial results from this work have been submitted for presentation at 2017 IEEE International Symposium on Biomedical Imaging (ISBI) 2018.}
\thanks{This work is accompanied with additional multimedia supplementary material. Images are best viewed in color on the electronic version of this document.}% <-this % stops a space

}

% note the % following the last \IEEEmembership and also \thanks - 
% these prevent an unwanted space from occurring between the last author name
% and the end of the author line. i.e., if you had this:
% 
% \author{....lastname \thanks{...} \thanks{...} }
%                     ^------------^------------^----Do not want these spaces!
%
% a space would be appended to the last name and could cause every name on that
% line to be shifted left slightly. This is one of those "LaTeX things". For
% instance, "\textbf{A} \textbf{B}" will typeset as "A B" not "AB". To get
% "AB" then you have to do: "\textbf{A}\textbf{B}"
% \thanks is no different in this regard, so shield the last } of each \thanks
% that ends a line with a % and do not let a space in before the next \thanks.
% Spaces after \IEEEmembership other than the last one are OK (and needed) as
% you are supposed to have spaces between the names. For what it is worth,
% this is a minor point as most people would not even notice if the said evil
% space somehow managed to creep in.

% The paper headers
\markboth{}%
{Shell \MakeLowercase{\textit{et al.}}: Bare Demo of IEEEtran.cls for Journals}
% The only time the second header will appear is for the odd numbered pages
% after the title page when using the twoside option.
% 
% *** Note that you probably will NOT want to include the author's ***
% *** name in the headers of peer review papers.                   ***
% You can use \ifCLASSOPTIONpeerreview for conditional compilation here if
% you desire.

% If you want to put a publisher's ID mark on the page you can do it like
% this:
%\IEEEpubid{0000--0000/00\$00.00~\copyright~2007 IEEE}
% Remember, if you use this you must call \IEEEpubidadjcol in the second
% column for its text to clear the IEEEpubid mark.

% use for special paper notices
%\IEEEspecialpapernotice{(Invited Paper)}

% make the title area
\maketitle

\begin{abstract}

Colorectal cancer is the fourth leading cause of cancer deaths worldwide and the second leading cause in the United States. The risk of colorectal cancer can be mitigated by the identification and removal of premalignant lesions through optical colonoscopy. Unfortunately, conventional colonoscopy misses more than 20\% of the polyps that should be removed, due in part to poor contrast of lesion topography. Imaging tissue topography during a colonoscopy is difficult because of the size constraints of the endoscope and the deforming mucosa. Most existing methods make geometric assumptions or incorporate \textit{a priori} information, which limits accuracy and sensitivity. In this paper, we present a method that avoids these restrictions, using a joint deep convolutional neural network-conditional random field (CNN-CRF) framework. Estimated depth is used to reconstruct the topography of the surface of the colon from a single image. We train the unary and pairwise potential functions of a CRF in a CNN on synthetic data, generated by developing an endoscope camera model and rendering over 100,000 images of an anatomically-realistic colon. We validate our approach with real endoscopy images from a porcine colon, transferred to a synthetic-like domain, with ground truth from registered computed tomography measurements. The CNN-CRF approach estimates depths with a relative error of 0.152 for synthetic endoscopy images and 0.242 for real endoscopy images. We show that the estimated depth maps can be used for reconstructing the topography of the mucosa from conventional colonoscopy images. This approach can easily be integrated into existing endoscopy systems and provides a foundation for improving computer-aided detection algorithms for detection, segmentation and classification of lesions.

\end{abstract}
\begin{IEEEkeywords}

\end{IEEEkeywords}

% For peer review papers, you can put extra information on the cover
% page as needed:
% \ifCLASSOPTIONpeerreview
% \begin{center} \bfseries EDICS Category: 3-BBND \end{center}
% \fi
%
% For peerreview papers, this IEEEtran command inserts a page break and
% creates the second title. It will be ignored for other modes.
\IEEEpeerreviewmaketitle

\hyphenation{op-tical net-works semi-conduc-tor}
\iffalse \bibliography{bibfile.bib} \fi
Endoscopy, Colonoscopy, Deep Learning, Conditional Random Fields, Monocular Depth Estimation, Learned Depth
\section{Introduction}
% The very first letter is a 2 line initial drop letter followed
% by the rest of the first word in caps.
% 
% form to use if the first word consists of a single letter:
% \IEEEPARstart{A}{demo} file is ....
% 
% form to use if you need the single drop letter followed by
% normal text (unknown if ever used by IEEE):
% \IEEEPARstart{A}{}demo file is ....
% 
% Some journals put the first two words in caps:
% \IEEEPARstart{T}{his demo} file is ....
% 
% Here we have the typical use of a "T" for an initial drop letter
% and "HIS" in caps to complete the first word.

% For peer review papers, you can put extra information on the cover
% page as needed:
% \ifCLASSOPTIONpeerreview
% \begin{center} \bfseries EDICS Category: 3-BBND \end{center}
% \fi
%
% For peerreview papers, this IEEEtran command inserts a page break and
% creates the second title. It will be ignored for other modes.
%\IEEEpeerreviewmaketitle
{}
% no \IEEEPARstart

%\IEEEPARstart{C}{olorectal} cancer is the fourth leading cause of cancer deaths worldwide \cite{winawer_history_2015,winawer_prevention_1993} and the second leading cause in the United States \cite{noauthor_colorectal_2012}. The detection and removal of premalignant lesions through an endoscopic colonoscopy is the most effective way to reduce colorectal cancer mortality. However, this approach has well-known limitations \cite{ransohoff_how_2009,baxter_association_2009} and recent studies have suggested that gastroenterologists can easily miss between 25\% of polyps \cite{leufkens2012factors,pabby2005analysis,van2006polyp,kim2007ct}. Around 60\% colorectal cancer cases detected after optical colonoscopy can be attributed to missed lesions \cite{heresbach2008miss}. Reconstructing the surface topography of the mucosa could help improve lesion detection and identification rates and could also aid in automating lesion detection. Current state-of-the-art lesion detection and classification methods rely on color and texture of the lesions for feature extraction \cite{iakovidis2005comparative,alexandre2008color}. However, the depth of the endoscopic scene holds information that could prove vital for lesion detection, classification as well as for topographical reconstructions which could be used to create a mosaic of the entire colon. Moreover, the 3D topographical information can be used to establish more accurate colonoscopy quality metrics.

\IEEEPARstart{C}OLORECTAL cancer (CRC) is the third most commonly diagnosed cancer in the United States \cite{siegel2017colorectal}. Colonoscopy screening can significantly reduce colorectal cancer mortality by detecting and removing premalignant lesions. However, this approach has well-known limitations \cite{ransohoff_how_2009,baxter_association_2009} and recent studies have suggested that gastroenterologists can easily miss more than 20\% of clinically relevant polyps \cite{leufkens2012factors,pabby2005analysis,van2006polyp,kim2007ct}. Approximately 60\% of colorectal cancer cases detected after optical colonoscopy are associated with missed lesions \cite{le2013postcolonoscopy,heresbach2008miss}. In addition to the well-characterized problem of missed polyps, non-polypoid lesions are even more difficult to screen for and are increasingly recognized as harboring significant malignant potential \cite{soetikno2008prevalence}. 

%In addition to the well-characterized problem of missed polyps, non-polypoid lesions are even more difficult to screen for and are increasingly recognized as harboring significant malignant potential.

One of the most effective ways to increase lesion detection rates is by using chromoendoscopy. Chromoendoscopy increases both polypoid and non-polypoid lesion contrast by iteratively spraying and rinsing a topical dye through the colon, effectively encoding surface topography as color contrast \cite{durr20143d}. However, chromoendoscopy is not used in routine screening because it doubles procedure time and requires specialized training.

Computational measurement of colon topography has the potential to improve lesion detection rates while meeting practical colonoscopy workflow and clinical constraints. Surface features could be used to amplify lesion contrast, assist in geometric lesion classification \cite{axon2005update}, augment conventional images \cite{gonzalez2014feature}, or improve computer-aided lesion detection algorithms. Current state-of-the-art lesion detection and classification methods rely on color and texture of the lesions for feature extraction \cite{iakovidis2005comparative,alexandre2008color}. However, surface topography of the colon could prove vital for automatic lesion detection and classification. Lastly, 3D mapping of the colon surface may be useful for advanced colonoscopy quality metrics, such as fractional coverage of the examination \cite{de2010advanced}.

\subsection{Related Work} % (fold)
\label{sub:related_work}

% subsection subsection_name (end){}

\textbf{Depth Estimation from Endoscopy Images} 

%Modern colonoscopes provide high definition wide angle videos but do not provide 3D reconstructions or topographical information of the surface of the colon. Despite the recent advancements in computer vision (CV) and image processing, optical colonoscopy remains a highly challenging environment specifically for depth estimation and 3D reconstruction. Such reconstructions from colonoscopy video or images is particularly challenging because endoscopes have a monocular camera and the movement of the endoscope is variable. The unpredictable movement, limited working area, limited endoscope size, non-uniform colon texture patterns and deformable nature of the colon render standard CV techniques like shape from stereo (SfSt) \cite{cohen_hierarchical_1989} and shape form texture (SfT) \cite{aloimonos_shape_1988,lobay_shape_2006} inadequate. Over the years, these has been some effort to reconstruct the surface of the colon from monocular images such as \cite{hong_3d_2014}, but such approaches have used unrealistic assumptions such as assuming the colon to be a tube etc.

Despite the recent advances in computer vision (CV) and image processing, colonoscopy remains a particularly challenging environment for depth estimation and 3D reconstruction. Colonoscopes have a monocular camera with close light sources, a wide field-of-view, and both the endoscope and the colon are in frequent motion. The unpredictable movement, limited working area, small endoscope size, non-uniform colon texture patterns and deformable nature of the colon render conventional CV techniques like shape from stereo (SfSt) \cite{cohen_hierarchical_1989} and shape from texture (SfT) \cite{aloimonos_shape_1988,lobay_shape_2006} inadequate for robust depth estimation. More advanced approaches attempt to reconstruct colon surfaces with restrictive assumptions \cite{hong_3d_2014}, and there is currently no model-based approach that robustly and accurately estimates depth from a colonoscopy video. Photometric stereo endoscopy captures some 3D structure of the mucosa \cite{durr2014system,durr20143d,parot2013photometric} but is inherently qualitative due to the unknown working distances from each object point to the endoscope.

\textbf{Learning for Depth Estimation} 

%Learning based methods have been used for monocular depth prediction for mainstream vision \cite{ranftl2016dense,li2015depth,eigen2014depth,liu_learning_2016,saxena20083,saxena2006learning} particularly for autonomous navigation. A dictionary learning-based approach has been employed for depth estimation for colonoscopy images in \cite{tourassi_computer-aided_2016}. However, the virtual colonoscopy data used for training does not follow optical properties of an actual endoscope and there is no validation of the results presented on real images. Preliminary results have been shown for 3D monocular reconstruction from ENT in \cite{reiter2016endoscopic}. However, the network used was shallow with a single layer and eight nodes trained from only 36 images. A small number of images were used because it is difficult to get ground true training data. Moreover, the texture in the data is patient specific and can not be used to estimate depth from other patients. Recent work on monocular 3D reconstruction for assisted navigation in bronchoscopy presented in \cite{visentini2017deep} uses deep learning for monocular depth estimation. However, the bronchoscopy environment lacks the presence of ridges and arbitrary motion of the colon which are problems in colonscopy. Moreover, the system is not tested and validated on real data.

Learning-based methods have been used for monocular depth prediction for conventional computer vision applications \cite{ranftl2016dense,li2015depth,eigen2014depth,liu_learning_2016,saxena20083,saxena2006learning}, particularly for autonomous navigation \cite{michels2005high,royer2007monocular}. A dictionary learning-based approach has recently been employed for depth estimation for colonoscopy images in \cite{tourassi_computer-aided_2016}. However, the virtual colonoscopy data used for training does not simulate the optical properties of an actual endoscope and there was no validation of the technique presented on real endoscopy images. Promising preliminary results have also been shown for 3D monocular reconstruction for functional endoscopic sinus surgery \cite{reiter2016endoscopic}, but the network used was shallow with a single layer and eight nodes trained from only 36 images. A small number of images were used because it is difficult to get ground truth training data. 
Moreover, the texture in the data is patient specific and cannot be used to estimate depth from other patients, requiring a new training set to be acquired at the beginning of each new procedure. Recent work on monocular 3D reconstruction for assisted navigation in bronchoscopy uses deep learning for monocular depth estimation \cite{visentini2017deep}. However, they validate their method only on phantom data and require training from patient specific CT data every time depth estimates are required from a new patient.

%However, the bronchoscopy environment lacks the presence of ridges and arbitrary motion of the colon which are problems in colonoscopy. 
%Moreover, this work required a new training set to be acquired at the beginning of each new procedure due to the uniqueness of the high-spatial frequency texture of each patient. %Moreover, the system is not tested and validated on real data.

%More recently, the concept of combining DCNNs with a graphical model for structured learning problems has set new records for a variety of CV tasks . Pixel-wise labeling, that assigns a continuous or discrete label to each pixel of the image, has generally been tackled with with feature engineering. Recently DCNNs have been extensively used with great success for a variety of CV tasks . However, such approaches lack spatial consistency such as smooth transitions etc. Spatial consistency has traditionally been captured by probabilistic graphical models such as CRFs \. Therefore a combination of DCNNs and CRFs is a natural augmentation of existing DCNN frameworks. For mainstream vision this formulation has been used for depth estimation in . 

The concept of combining CNNs with a graphical model for structured learning problems has demonstrated considerable promising results \cite{liu_learning_2016,eigen_depth_2014,schmidhuber_deep_2015}. Pixel-wise labeling, that assigns a continuous or discrete label to each pixel of the image, has generally been tackled with feature engineering. Recently CNNs have been extensively used with great success for a variety of CV tasks \cite{schmidhuber_deep_2015,lecun_deep_2015}. However, such approaches lack spatial consistency such as smooth transitions. Spatial consistency has traditionally been captured by probabilistic graphical models such as CRFs \cite{lafferty_conditional_2001}, and can augment CNNs to improve depth estimation accuracy \cite{liu_learning_2016,li2015depth}.

\subsection{Contributions and Significance} % (fold)
\label{sub:contributions}

%Traditional approaches of learning-based depth estimation have relied on features from texture or color to estimate the depth. We propose that the strongest cue for depth estimation in an endoscopic scene is the inverse of intensity fall-off. Most of the texture inside the colon is patient specific, a fact that has limited previous learning-based approaches to work on a diverse set of test data. Most virtual colonoscopy software such as Slicer 3D \cite{nain2001interactive} and Viatronix \cite{pickhardt2003computed} does not present an accurate model for the optical properties of an endoscope. In this work we train a CNN-CRF network on optically accurate endoscopy data with ground true depth. Our main contributions can be summarized as follows: 

Traditional approaches of learning-based depth estimation are trained on data that include patient-specific texture, color, and shape, making them difficult to generalize without acquiring a large amount of ground truth data. Low-level texture details are patient-specific and not diagnostic, such as vascular patterns. High-level texture, on the other hand, contains clinically-relevant features that can be generalized across patients. Although details from texture and color are important, these approaches fail to exploit what may be the strongest cue of depth in the small working distances encountered in endoscopy, the inverse square fall-off in light intensity with propagation distance. Most CT colonoscopy (CTC) or virtual colonoscopy software packages such as Slicer 3D \cite{fedorov20123d,nain2001interactive} and Viatronix \cite{pickhardt2003computed} do not utilize an accurate model for the optical properties of an endoscope. 

Depth values for a specific object in view of the endoscope are inherently continuous, thus depth estimation from monocular endoscope images can easily be formulated as a Conditional Random Fields (CRF) learning problem. In this work we develop and train a CNN-CRF network on a large dataset of realistic endoscopy data with ground truth depth. Our specific contributions can be summarized as follows:

\begin{enumerate}
\item We developed an accurate optical model of an endoscope that includes the inverse-square law of intensity fall-off, to generate synthetic and virtual images of the colon with ground truth depth. 
%This virtual endoscope is used to generate a large database of images with corresponding ground truth depth from a synthetically-designed, texture-free colon with realistic size and shape. The virtual endoscope is also used to render images from a CT-scan of a silicone colon phantom that is imaged with a real optical endoscope. Images from both approaches are registered to create a dataset of real endoscope images with corresponding depth maps.
\item We use this data to train a network that consists of the unary and pairwise parts of a joint CNN-CRF framework.
%\item We use adversarial training to train a transformation network which transforms test images to their synthetic-like counterpart for domain adaptation.
\item We validate our results on test data from the digital synthetic colon, a silicone colon phantom, and endoscopy images collected from a porcine colon registered with CT to give accurate ground truth depth.
\end{enumerate}

%To the best of our knowledge this is the first work where a deep learning network is trained on synthetic data for a medical imaging problem. With deep learning setting records for a variety of applications, for these advantages to trickle down to the medical imaging domain large amounts of data is required for training. Unfortunately, large amounts of labeled or annotated medical data is not readily available due to privacy issues, scarcity of experts available for annotation, underrepresentation of rare conditions which leads to highly correlated features of the normal condition and non standardized datasets etc. All these issues with generation of ground true medical data can be addressed if an accurate forward model for the imaging system and an anatomically accurate model of the organ can be computationally built. In this work we demonstrate that it is possible to train a network on synthetically rendered data.

%With deep learning demonstrating considerable promise for a variety of image analysis applications, this general concept may be valuable for translating the benefits of deep learning to medical imaging. 
To the best of our knowledge, this is the first deep learning network trained on synthetically-generated endoscopy images. In practice, large datasets of labeled or annotated medical images are not generally available due to privacy issues, scarcity of experts available for annotation, underrepresentation of rare conditions which leads to highly correlated features of the normal condition and non standardized datasets. This problem has recently been tackled with transfer learning, which shows promising results on conventional computer vision networks fine tuned for medical images \cite{ravishankar2016understanding,litjens2017survey,tajbakhsh2016convolutional,yap2017automated}. However, transfer learning can lead to artifacts specifically for regression problems \cite{ravishankar2016understanding}. We show that the significant performance benefits of training with large datasets can be realized by utilizing synthetically-generated medical data with an accurate forward model for the imaging system and an anatomically-realistic model of the organ. We further show that this network accurately estimates depth in real endoscopy images after transfer to a synthetic-like domain. 
%\fm{For adapting our model trained on synthetic images to real endoscopy images we use a transformer network trained via adversarial training.}
%This is also and also the first work that combines DCNNs and CRFs for depth estimation from endoscopy images. Endoscopy is well-suited for this approach because it does not require any knowledge of object geometry, depth map continuous ….he first work that combines DCNNs and CRFs for depth estimation from single endoscopy images. \

\begin{figure}
\centering
\includegraphics[width=8.5cm]{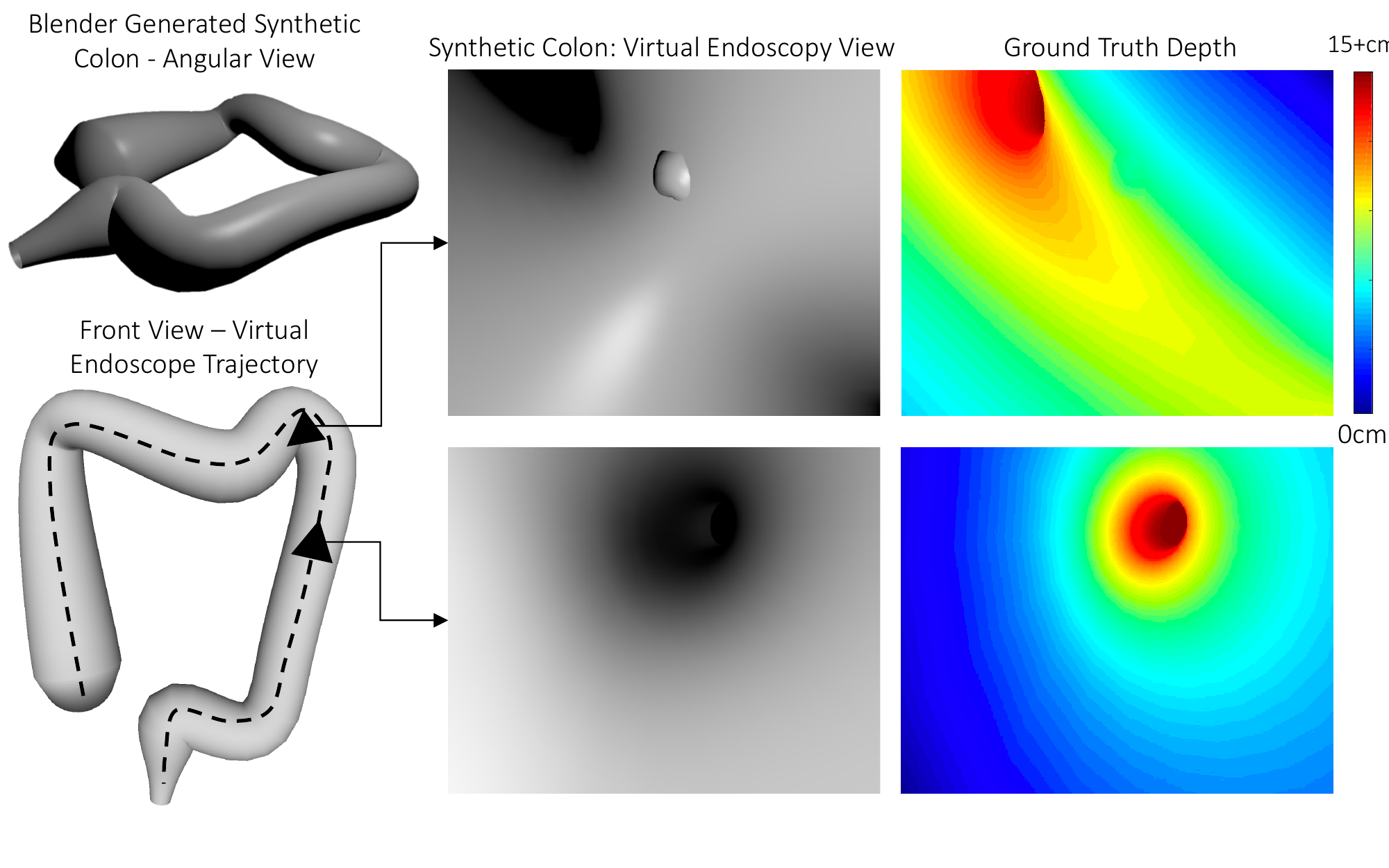}
\caption{Training frames and ground truth depths from a texture free synthetic colon. This model was used to generate a large dataset of rendered images with low spatial frequency topography and noise-free ground-truth depth.}
\label{fig_sim}
\vspace{-4.8mm}
\end{figure}

\begin{figure}
\centering
\includegraphics[width=8.5cm]{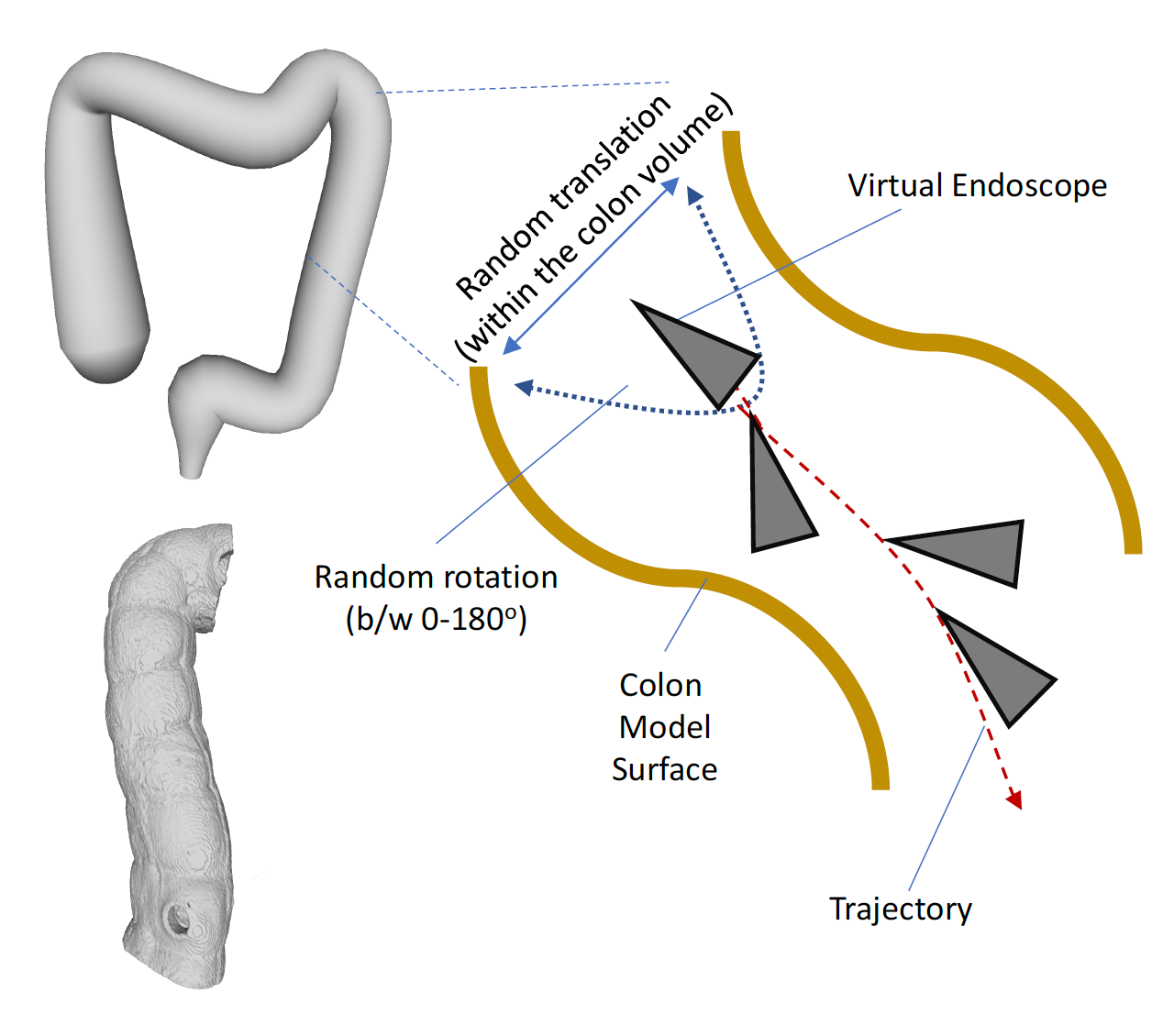}
\caption{Digital data collection using a virtual endoscope moving through a synthetic colon. The endoscope was randomly translated and rotated to generate a variety of data with ground truth depth for training.}
\vspace{-3.8mm}
\end{figure}

\section{Methods}

\subsection{Generating Ground Truth Training and Test Data}

 %A significant amount of endoscopy data with known depth maps is required for training a CNN for learning depth from a monocular scene. To the best of our knowledge there are no available optical endoscopy datasets with an accurate frame-by-frame depth recorded. Obtaining such a dataset poses a significant challenge since it is impractical to have a depth sensor coupled with the endoscope during a colonoscopy procedure. Moreover, the texture of the colon is patient-specific and cannot be used to efficiently learn depth. 

A large dataset of endoscopy images with corresponding ground truth depth maps is required for training a CNN to estimate depth from a monocular scene. This data is challenging to generate because depth sensors are impractical to couple to a small endoscope and must receive regulatory approval to be used in humans. Moreover, the high level texture of the colon is patient-specific and cannot be used to efficiently learn depth. To circumvent these obstacles, we generate several datasets of images from synthetic, phantom, porcine, and human models, which have, increasing realism, decreasing quality of ground truth depth, and decreasing dataset sizes.

 \textbf{Synthetic Colon - Virtual Endoscopy Data}.

%To train our network, we generated almost 100,000 texture free endoscopy images, each with an associated ground true depth from a virtual colon phantom. This virtual colon phantom was generated using Blender \cite{hess2007essential} and the data was recorded using a virtual endoscope with parameters, viewing angle, saturation etc. adjusted to mimic a real endoscope. The virtual colon had anatomically comparable diameter, bending angles and ployp sises (Fig. 1). The rendered images have a resolution of $720\times576px$ and a varying viewing angle between $110-140^{o}$. A Mitchell-Netravalie filter \cite{mitchell1988reconstruction,lehmann1999survey} was used to prevent aliasing. Two virtual light sources were coupled with the virtual endoscope. These point sources had an inverse square fall-off varying between 10-15cm. The depth of the scene was recorded by calculating the distance from the virtual endoscope to each point on the synthetic colon being imaged. Fig. 1 shows the synthetic colon and some endoscopy data with aligned depth generated using this procedure. In order to accurately model the effects of endoscope motion and light falling on similar surfaces from different angles the virtual endoscope was rotated at a variety of angles to collect a diverse set of data. Instead of traditional data augmentation we changed the properties of our virtual endoscope to generate a diverse set of endoscopy data.

To train our network, we generated over 100,000 texture free endoscopy images, each with an associated ground truth depth from a digital synthetic colon. This synthetic colon phantom was generated using Blender \cite{hess2007essential} and the data were recorded using a virtual endoscope with parameters selected to mimic the range found in common colonoscopes. The virtual colon had anatomically realistic diameter, bending angles and polyps \cite{hounnou2002anatomical} (Fig. 1). The rendered images have a resolution of $720\times576px$ and a varying viewing angle between $110-140^{o}$. A Mitchell-Netravalie filter \cite{mitchell1988reconstruction,lehmann1999survey} was used to prevent aliasing. Two virtual light sources were placed on either side of the camera on the virtual endoscope and each was configured to provide inverse square fall-off of illumination intensity. The depth of the scene was recorded by calculating the distance from the camera to each point on the synthetic colon being imaged. Fig. 1 shows the synthetic colon and representative endoscopy data with aligned depth maps generated using this procedure. We varied the position of the virtual endoscope to generate a diverse set of endoscopy data and in order to accurately model the effects of endoscope motion and light illuminating similar surfaces from different angles (Fig 2). The virtual endoscope was randomly translated along the horizontal axis within the bounds of the synthetic colon and randomly rotated between $0-180^{0}$ to generate a diverse set of data. This dataset was used for pre-training the CNN-CRF network.

\begin{figure}
\centering
\includegraphics[width=8.5cm]{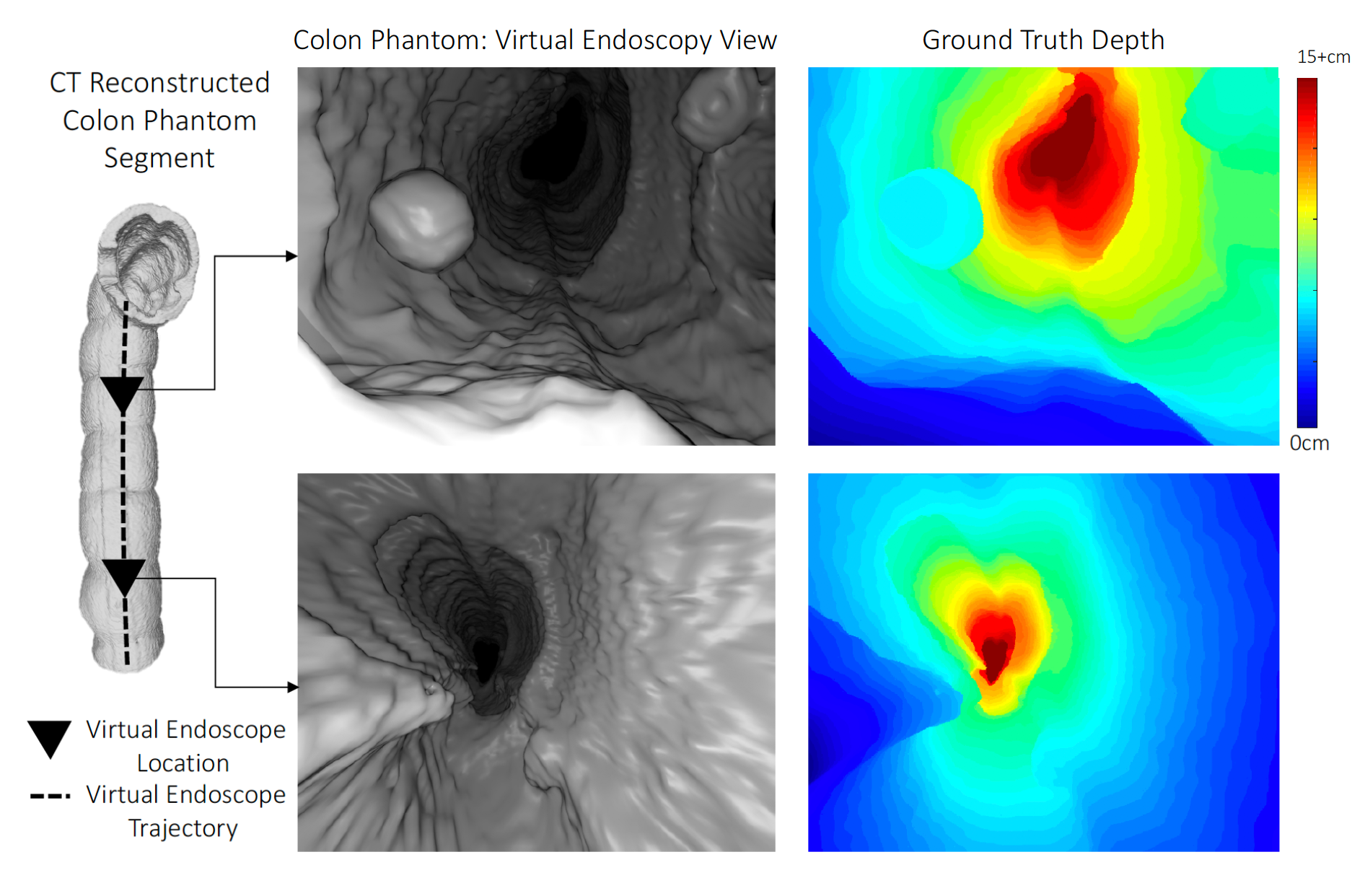}
\caption{Training frames and ground truth depths from reconstructed CT of a silicone colon phantom model. The 3D model was imaged using our virtual endoscope to generate training data with realistic high-spatial frequency topography.}
\label{fig_sim}
\vspace{-3.8mm}
\end{figure}

\begin{table*}
 \captionsetup{justification=centering,   textfont={sc}}
  \caption{A Comparison of different endoscopy datasets.}

\centering
\begin{tabular}{|c|c|c|c|c|c|c|c|}
\hline
%\rowcolor{gray}
\textbf{Dataset} & \textbf{\thead{Virtual /\\Real Endoscopy}} & \textbf{\thead{Low Spatial \\ Freq. Detail}} & \textbf{\thead{High Spatial \\ Freq. Detail}} & \textbf{\thead{Ground Truth \\ Depth Available}} & \textbf{\thead{Training}} & \textbf{Testing} & \textbf{Dataset Size}\\    \hline
\makecell{\textbf{Synthetic Colon}}	& Virtual & Yes 			& No   & Yes & \makecell{Yes} 		& Yes &100,000\\    	 \hline
\makecell{\textbf{Phantom Colon}}		& Virtual & Yes 			& Yes  & Yes & \makecell{Yes} 		&Yes &100,000\\   	 \hline
\makecell{\textbf{Porcine Colon}}			& Real & Yes 			& Yes  & Yes & \makecell{No} 		& Yes&1460\\   	 \hline
\makecell{\textbf{Human Colonoscopy}}		& Real & Yes 			& Yes  & No  & \makecell{No} 		& Qualitative& N/A\\    	 \hline \hline
\makecell{\textbf{Human CTC}}			& Virtual & Yes & No  & Yes & \makecell{Not Used}  & Not Used & N/A\\ \hline

\end{tabular}
\end{table*}

\textbf{Colon Phantom CT - Virtual Endoscopy Data}.

%Although the synthetically rendered texture free endoscopy data is sufficient for learning the inverse of intensity it does lack some high frequency information. We also train our network with virtual endoscopy data collected from a reconstructed CT of a colon phantom (The Chamberlain Group, Colonoscopy Trainer \#2003). The CT reconstruction was performed using filtered back-projection and was filtered using a ram-lak filter in Slicer 3D. The data was then exported to blender where it was imaged using the same 
%virtual endoscope described in the prequel. Such endoscopy images help the network learn the properties of inverse intensity fall-off on ridges and polyps in the colon. This CT reconstructed data is deliberately smoothened to reduce the effects of any fine texture in the scene which may be specific to the patient the phantom was molded from. Almost 100,000 images are collected from this setup including augmented images as described previously. Fig. 2 shows the reconstructed colon phantom and rendered virtual endoscopy images and aligned ground true depth.

Although the synthetic colon data may be sufficient for learning the inverse square law, our synthetic colon model lacks high spatial frequency detail. To incorporate sensitivity to these features, we also generated training data from virtual endoscopy images from a CT dataset of a silicone colon phantom (The Chamberlain Group Colonoscopy Trainer \#2003). The CT reconstruction was performed using filtered back-projection and was filtered using a Ram-Lak filter with linear interpolation in Slicer 3D \cite{fedorov20123d}. The data was then imaged using the virtual endoscope described previously. These endoscopy images with higher frequency details help the network learn the properties of inverse intensity fall-off on ridges and polyps in the colon. This reconstructed CT data was filtered to reduce the effects of fine texture in the scene which may be specific to the cadaver the phantom was molded from. Over 100,000 images were collected from this setup. Fig. 3 shows a portion of the reconstructed colon phantom, rendered virtual endoscopy images, and corresponding ground truth depth. This dataset was used to fine-tune our CNN-CRF depth estimation network.

\textbf{Porcine Colon CT - Real Endoscopy Data}.

\begin{figure*}
\centering
\includegraphics[width=\textwidth]{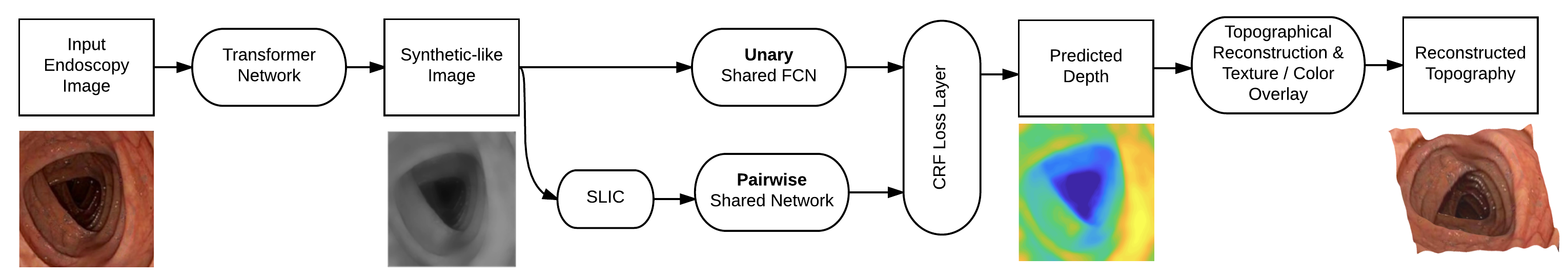}
\caption{Monocular endoscopy depth prediction framework. The input image is transformed to its synthetic-like representation and fed into a shared unary and pairwise network trained on synthetic endoscopy data. The unary part regress the depth and the pairwise part is responsible for smoothing based on neighboring superpixels. The predicted depth can be used to reconstruct a topographical map of the surface of the mucosa. }
\label{fig_sim}
\vspace{-1.4em}
\end{figure*}

%To validate the effectiveness of our trained models on real tissue we dissected a porcine colon and placed it on a scaffold with an anatomically accurate bend (Fig. 5). The scaffold was then imaged using a CT scanner and 720 projections were collected, each at half a degree, rotating the scaffold on a stepper motor stage. The scaffold was then covered with black foil and was images with a Misumi (MO-V5006L) wide angle endoscope. The CT projects were reconstructed using filtered back-projection and a ram-lack filter. The resulting 3D model was imaged using our virtual endoscope in blender. The CT and optical endoscopy results were registered using control point and feature detection and matching. This setup has been shown in Fig. 6. \\
%To the best of our knowledge this is the first real endoscopy dataset 

To validate the accuracy of our trained model on real tissue, we dissected a porcine colon and fixed it to a half-pipe scaffold with a diameter of $5.1cm$ and a $90$ degree bend to simulate the anatomy of the human transverse and descending colon (Fig. 7). Metallic pins were used as fiducial markers for localization and size estimation. The tissue was then imaged using a benchtop cone beam CT scanner with 720 projections at half-degree increments by rotating the scaffold on a stepper motor stage. The CT projections were reconstructed using filtered back-projection and a Ram-Lak filter \cite{kak2001principles}. The resulting 3D model was imaged using the virtual endoscope in Blender. The scaffold was then covered with black foil and was imaged with an optical endoscope (Misumi MO-V5006L) with a $104^{o}$ $2.3mm/f10$ wide angle lens with (Misumi L23010IR-M5.5-53). The CT and optical endoscopy results were registered to get ground truth depth maps.

%\textbf{Registering Virtual and Optical Endoscopy Data}

Optimization-based multimodal registration was performed on the virtual endoscopy image collected from a CT reconstructed porcine colon and an optical endoscopy image of the same view. A one-plus-one evolutionary optimizer \cite{styner2000parametric,styner1997evaluation} was used to optimize a Mattes mutual information metric for similarity. The growth factor was set to 1.52, initial radius was set to 0.30, the radius adjustment was set to 0.014 and the maximum number of random spatial samples used to compute the metric was  set to 500. The optimization was run for 250 iterations at 3 pyramid levels.

%\lipsum[16-50]

\textbf{Human Endoscopy Data}.
Human endoscopy data available from the NIH and other datasets available from the MICCAI endoscopy challenge \cite{tajbakhsh2016automated} were used to qualitatively evaluate the performance of our network. Since real endoscopy data can have specular reflections we used graph-based in-painting \cite{peyre2008non,liu2013exemplar} to partially remove specular reflections.\\

%A summary of the properties of these datasets has been given in Table I. 

Table I compares various datasets used in this study and CTC or virtual colonoscopy data. CTC has been used in other depth estimation studies such as \cite{tourassi_computer-aided_2016}, however poor cleaning is a major limiting factor \cite{pickhardt2003electronic}. Moreover, recent studies on nonpolypoid lesion detection using CTC restrict to polyps larger than $6$ mm because of the limited practical resolution of in-vivo CTC and the requirement of post-processing to remove artifacts from incomplete preparation \cite{pickhardt2004flat}. CTC also has a relatively high miss rate for non-polypoid lesions \cite{fidler2002}. For these reasons CTC data is not used for training.

\subsection{Deep Learning with Conditional Random Fields}

%The aim for this setup is to predict the depth from a monocular endoscopic image. Inspired by success of using similar models for real world images in \cite{qin_global_2009,xu2017multi,radosavljevic2010continuous,liu_learning_2016} we propose using continuous conditional random fields and DCNN in a joint framework presented below. Since depth values of a specific region in an endoscopic scene are inherently continuous using continuous CRFs has a specific advantage. Moreover, unlike several previous methods we do not need to make any assumptions since the log-likelihood optimization problem can be directly solved because the partition function can be analytically calculated. 
Inspired by success of using similar models for analyzing conventional images in \cite{qin_global_2009,xu2017multi,radosavljevic2010continuous,liu_learning_2016}, we implemented an algorithm to estimate depth using continuous CRF and CNN. A continuous CRF is able to exploit the continuous depth values within specific regions of an endoscopic scene. Moreover, unlike several previous methods, this method does not require assumptions since the log-likelihood optimization problem can be directly solved because the partition function can be analytically calculated. Fig. 4 shows a top level flow diagram of the setup. The following sections describe the unary and pairwise parts and the training in detail. Then we discuss how the network trained on virtual endoscopy data is adapted to real data.

\textbf{Preliminaries.} Let $\vec{x} \in \R^{n\times m}$ be an image acquired from an endoscopic camera which has been divided into $g$ super pixels and $\vec{y}=[y_1,y_2,...,y_g] \in \R^{g}$ be the depth vector corresponding each super-pixel. Based on conventional graphical models, the depth of an image can be predicted by solving the following maximum a postereori (MAP) problem:  

\begin{equation}\label{eq:Tk}
  \begin{aligned}
\widehat{\vec{y}}=\argmax_{y} {Pr(\vec{y}|\vec{x})}.
  \end{aligned}
\end{equation}

As with general CRFs the conditional probability distribution of the raw data can be defined as:

\begin{equation}\label{eq:Tk}
  \begin{aligned}
 {Pr(\vec{y}|\vec{x})}=\frac{exp(E(\vec{y},\vec{x}))}{\int_{-\infty}^{\infty} exp(E(\vec{y},\vec{x})) d\vec{y}}.
  \end{aligned}
\end{equation}

\begin{figure*}
\centering
\includegraphics[width=\textwidth]{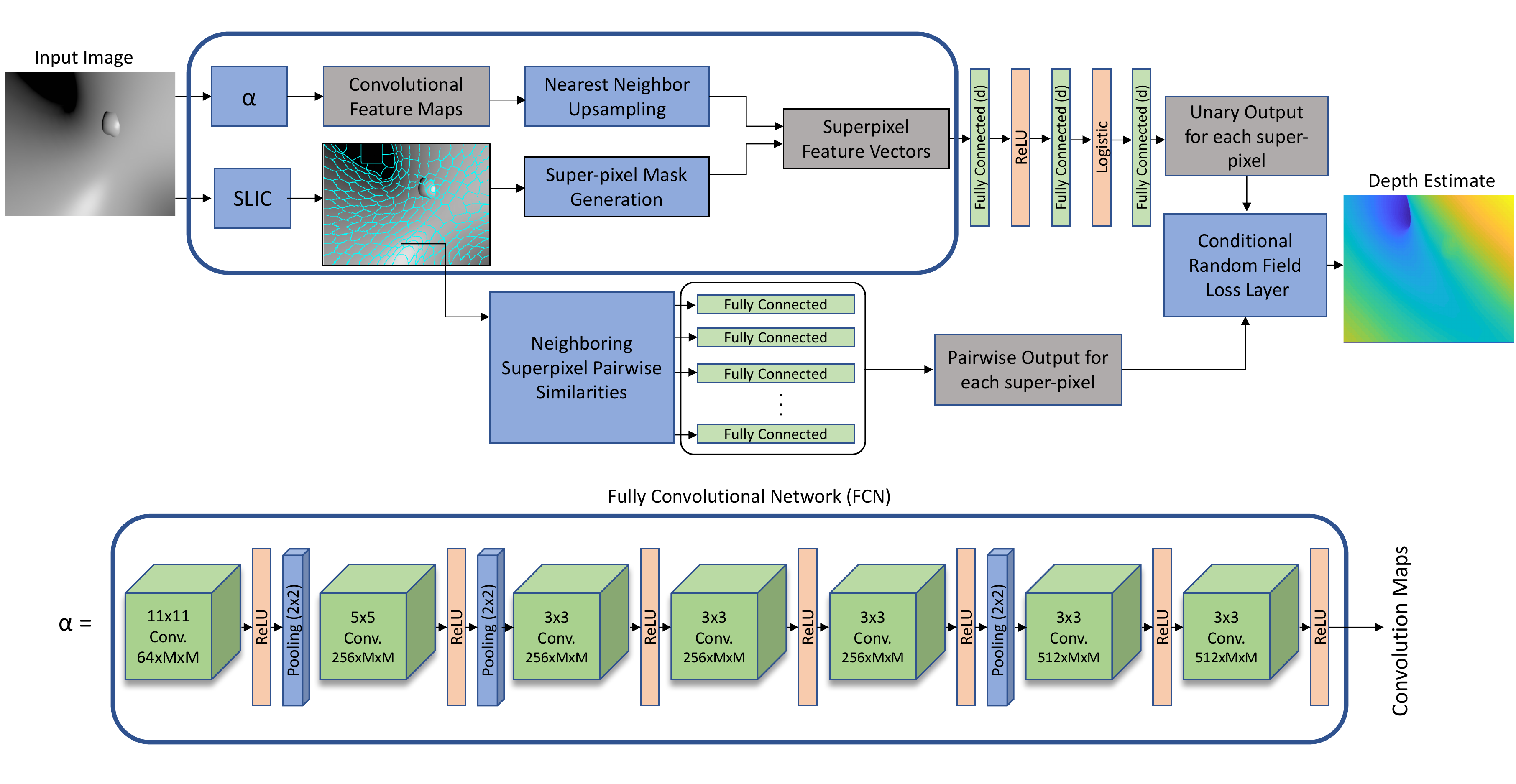}
\caption{The overall architecture for depth estimation from monocular endoscopy images. The input image is fed into a fully convolution network ($\alpha$) which produces convolution maps. The maps were then related back to super-pixels in a pooling layer which gives feature vectors for each super-pixel. This is followed by 3 fully connected layers which gives the unary part. For the pairwise part, similarities based on neighboring super-pixels were considered and fed into a fully connected layer. The outputs of both the unary and pairwise parts were fed into a CRF loss layer which minimizes the negative log likelihood of the probability density function.}
\label{fig_sim}
\vspace{-4mm}
\end{figure*}
 %Comparison of our proposed depth estimation and reconstruction method with a reconstruction from \cite{hong_3d_2014}. Input image and 3D reconstruction was taken from \cite{hong_3d_2014}.

 The energy function, $E(\vec{y},\vec{x})$ can be defined in terms of the unary potentials $\psi$ and pairwise potentials $\phi$ over nodes $\N$ and edges $\Sp$ of $\vec{x}$,

\begin{equation}\label{eq:Tk}
  \begin{aligned}
 E(\vec{y},\vec{x}) = \sum_{i \in \N} \psi({y}_i,\vec{x};\vec\gamma) + \sum_{(i,j) \in \Sp} \phi({y}_{i},{y}_{j},\vec{x};\vec\beta),
  \end{aligned}
\end{equation}

 where, the unary part, $\psi$, regresses the depth from each superpixel and the pairwise part, $\phi$, enforces smoothness between similar neighboring superpixels. $\gamma$ and $\beta$ are the two learning parameters associated with the unary and pairwise terms respectively.

 \textbf{Unary Potential.} The unary part is designed to regresses superpixel-wise depth for an input endoscopy image. Similar to \cite{liu_learning_2016} the unary potential can be defined as follows, 

\begin{equation}\label{eq:Tk}
  \begin{aligned}
 \psi(y_{i},\vec{x};\vec\gamma)= - (y_i-h_i(\vec{\gamma}))^2,
  \end{aligned}
\end{equation}

 where $h_i$ is the regressed depth of superpixel $i$ and $\gamma$ represents CNN parameters. The architecture of the training network is described systematically in Fig. 5. The CNN used for the unary part makes use of recent developments in fully convolutional networks (FCNs). Unlike standard CNNs which are composed of convolution followed by fully connected layers and produce non-spatial outputs, FCNs can take images of any size and produce spatial convolutional maps. FCNs have have been extensively used for complicated problems specifically for semantic segmentation \cite{eigen2014depth,long2015fully,dong2014learning,kamnitsas2017efficient}. We initialize the first five layers from Alex-Net \cite{krizhevsky2012imagenet,chatfield2014return}. Two additional $512$ channel convolutional layers with a filter size of $3\times 3$ are added to the network (as shown in $\alpha$, Fig. 5). $\alpha$ is capable of taking an input image of any size and giving $512$ channel convolutional maps. A typical problem with all fully convolutional architectures is that the feature maps produced can be significantly smaller than the actual size of the images. We mitigate this problem using a convolution map up-sampling step. For ease of implementation we use nearest neighbor up-sampling. Moreover, we incorporate a super-pixel pooling layer similar to \cite{kwak2017weakly,liu_learning_2016} to acquire super-pixel features from convolutional maps.

 \textbf{Pairwise Potential.} The objective of the pairwise potential is to smooth the depth regressed from the unary part based on the neighboring super-pixels. The pairwise potential function is based on standard CRF vertex and edge feature functions studied extensively in \cite{qin_global_2009} and other works. Let $\vec{\beta}$ be the network parameters and $\vec{S}$ be the similarity matrix where ${S}_{i,j}^k$ represents a similarity metric between ithe $i^{th}$ and $j^{th}$ superpixel. We hypothesize that intensity is a valuable cue for depth estimation. With this in mind, we use intensity difference and the greyscale histogram as pairwise similarities expressed in the general $\ell_2$ form. The pairwise potential can then be defined as,
\vspace{-4mm}
\begin{equation}\label{eq:Tk}
  \begin{aligned}
 \phi(y_{i},y_{j};\vec\beta)= - \frac{1}{2}\sum_{k=1}^{K}\beta_{k}S_{i,j}^{k}(y_i-y_j)^2.
  \end{aligned}
\end{equation}
\vspace{-4mm}

 %where $R_{pq}$ is the output of the pairwise fully connected network and $\beta$ represents a vector of network parameters. The pairwise shared network encourages smoothing based on neighboring pixels. 

 \textbf{Learning $h(\vec\gamma)$ and $\vec\beta$.} The overall energy function defined in Eq. 3 can now be populated with unary and pairwise terms and can be written as,
%\vspace{-0.7em}
\begin{equation}\label{eq:Tk}
  \begin{aligned}
 E = - \sum_{i \in \N} (y_i-h_i(\vec{\gamma}))^2  - \frac{1}{2}\sum_{(i,j) \in \Sp}\sum_{k=1}^{K}\beta_{k}S_{i,j}^{k}(y_i-y_j)^2.
  \end{aligned}
\end{equation}

For simplicity and explicit vector calculations the term $A_{i,j}=\sum_{k=1}^{K}\beta_{k}S_{i,j}^{k}$ can be defined as the affinity matrix, and $D_{i,i}=\sum_{j}A_{i,j}$ as a $n\times n$ diagonal matrix. $\vec{L}=\vec{D}-\vec{A}$ defines the graph Laplacian for further simplicity we notate $\vec{\xi}=\vec{I}+\vec{L}$ where $\vec{I}$ is a $n\times n$ identity matrix. Using these notations Eq. 6 can be simplified as,

\begin{equation}\label{eq:Tk}
  \begin{aligned}
 E = \vec{y}(2\vec{h}^{\top}-\vec{\xi}\vec{y}^{\top})-\vec{h}\vec{h}^{\top}.
  \end{aligned}
\end{equation}

Assigning $\vec{\zeta}=2\vec{h}^{\top}-\vec{\xi}\vec{y}^{\top}$, the probability density function in Eq. 2 can now be simplified to the following form,

\begin{equation}\label{eq:Tk}
  \begin{aligned}
 {Pr(\vec{y}|\vec{x})}=\sqrt{\frac{\vec{\lvert\xi\rvert}}{\pi^n}}exp\{{\vec{y}\vec{\zeta}-\vec{\xi}^{-1}\vec{h}\vec{h}^{\top}}\}
  \end{aligned}
\end{equation}

\begin{figure*}
\centering
\includegraphics[width=\textwidth]{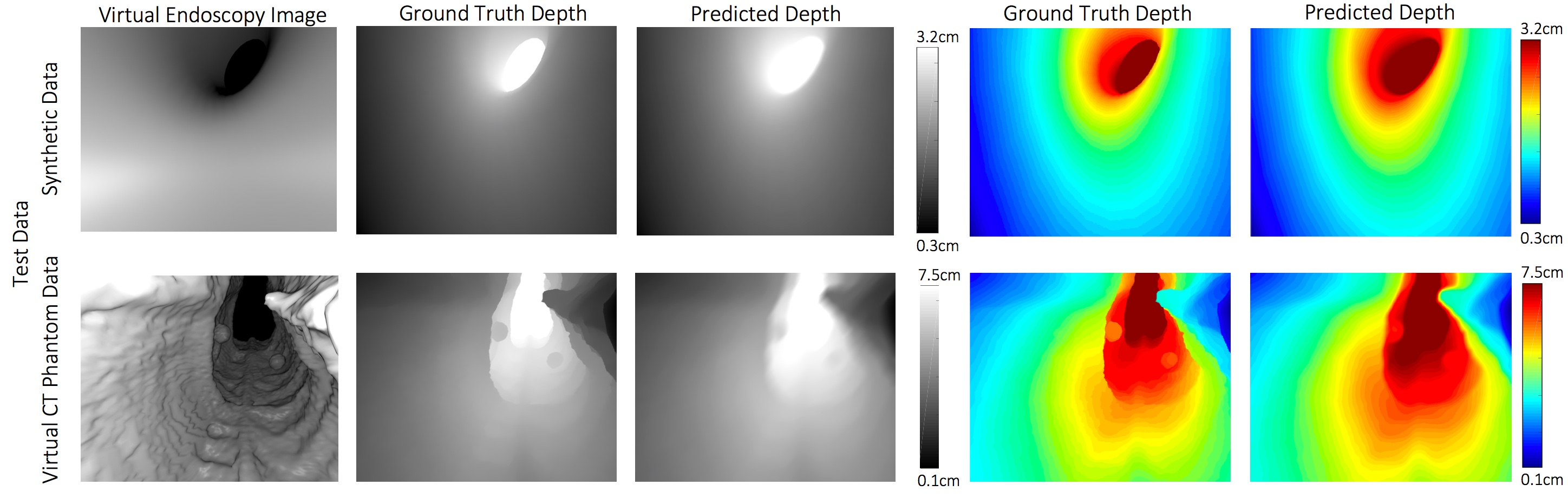}
%\vspace*{0.5mm}
\caption{Input frames and predicted depths for synthetic colon data and colon phantom CT data imaged using our virtual endoscope.}
\label{fig_sim}
\vspace*{-4.2mm}
\end{figure*}

Given $Pr(\vec{y}|\vec{x})$ we can now calculate the negative log-likelihood which simplifies to,

\begin{equation}\label{eq:Tk}
  \begin{aligned}
 {-\log Pr(\vec{y}|\vec{x})}={\vec{y}\vec{\zeta}-\vec{\xi}^{-1}\vec{h}\vec{h}^{\top}}-\frac{1}{2}\{\log\vec{\lvert\xi\rvert}+n\log(\pi)\}.
  \end{aligned}
\end{equation}

The negative log-likelihood of the training data is minimized during the training process and the optimization problem can be represented as the following objective function,

\begin{equation}\label{eq:Tk}
  \begin{aligned}
	\min_{\gamma,\beta \geq 0} {-\sum_{1}^{N}\log Pr(\vec{y}|\vec{x};\vec{\gamma},\vec{\beta}})+ \frac{\lambda_1}{2} \norm{\gamma}_2^2+ \frac{\lambda_2}{2} \norm{\beta}_2^2.
  \end{aligned}
\end{equation}

Where $N$ represents the maximum number of images in the training set. In order to prevent over-fitting two $\ell_2$ regularization terms have been added with each learning parameter. $\lambda_1$ and $\lambda_2$ represent the regularization or weight decay parameters. $\ell_2$ regularization penalizes heavily weighted vectors and promotes weight diffusion by encouraging the network to utilize all its inputs. 

\textbf{Optimization Solution.} The optimization problem is solved by standard stochastic gradient decent-based back-propagation. For the unary part the the partial derivatives of  $-\log Pr(\vec{y}|\vec{x})$ are calculated with respect to $\gamma$. In Eq. 9 only the terms with $\vec{h}$ represent a term with $\gamma$ so all other terms are excluded as a result of the partial derivative,

\begin{equation}\label{eq:Tk}
  \begin{aligned}
\frac{\partial \{-\log Pr(\vec{y}|\vec{x})\}}{\partial \gamma} = - 2\vec{y}^{\top} \frac{\partial \vec{h}}{\partial \gamma} + 2 \vec{h}^{\top}{\vec{\xi}^{-1}}\frac{\partial \vec{h}}{\partial \gamma}
  \end{aligned}
\end{equation}

For the pairwise part the partial derivatives are calculated with respect to $\beta$ 
%\small
\begin{equation}\label{eq:Tk}
  \begin{aligned}
\frac{\partial \{-\log Pr(\vec{y}|\vec{x})\}}{\partial \beta} = \vec{y}\vec{y}^{\top} \frac{\partial \vec{\xi}}{\partial \gamma} - \vec{h}\vec{h}^{\top} \vec{\xi}^{-1} (\vec{\xi}^{-1})^{\top} \frac{\partial \vec{\xi}}{\partial \gamma} \\ - \frac{1}{2} \frac{1}{\lvert \vec{\xi}\rvert} \frac{\partial \lvert \vec{\xi}\rvert}{\partial \gamma}
%- 2\vec{y}^{\top} \frac{\partial \vec{h}}{\partial \gamma} + 2 \vec{h}^{\top}{\vec{\xi}^{-1}}\frac{\partial \vec{h}}{\partial \gamma}
  \end{aligned}
\end{equation}
%\normal
%\notmalsize

\textbf{Depth Estimation.} To estimate the depth of a new endoscopic image, the MAP problem in Eq. 1 must be solved. Here we show that a closed form solution of the problem exists based on the definitions presented above. 

\begin{equation}\label{eq:Tk}
  \begin{aligned}
\widehat{\vec{y}}= \argmax_{y} \bigl\{{{\vec{y}\vec{\zeta}-\vec{\xi}^{-1}\vec{h}\vec{h}^{\top}}-\frac{1}{2}\{\log\vec{\lvert\xi\rvert}+n\log(\pi)\}}\bigr\}
  \end{aligned}
\end{equation}

To solve the above maximization problem, the partial derivative of the maximization term has to be calculated with respect to $\vec{y}$. Thus, all terms without $\vec{y}$ can be ignored, simplifying the problem to,

\begin{equation}\label{eq:Tk}
  \begin{aligned}
\frac{\partial \{\vec{y}(2\vec{h}^{\top}-\vec{\xi}\vec{y}^{\top})\}}{\partial \vec{y}} = 0 \implies -\vec{y}^2\vec{\xi}+2\vec{h}\vec{y} = 0 \\
\widehat{\vec{y}}=\vec{\xi}^{-1}\vec{h}
  \end{aligned}
\end{equation}

This clearly shows that the problem has a close form solution and can be solved.

\begin{figure*}
\centering
\includegraphics[width=\textwidth]{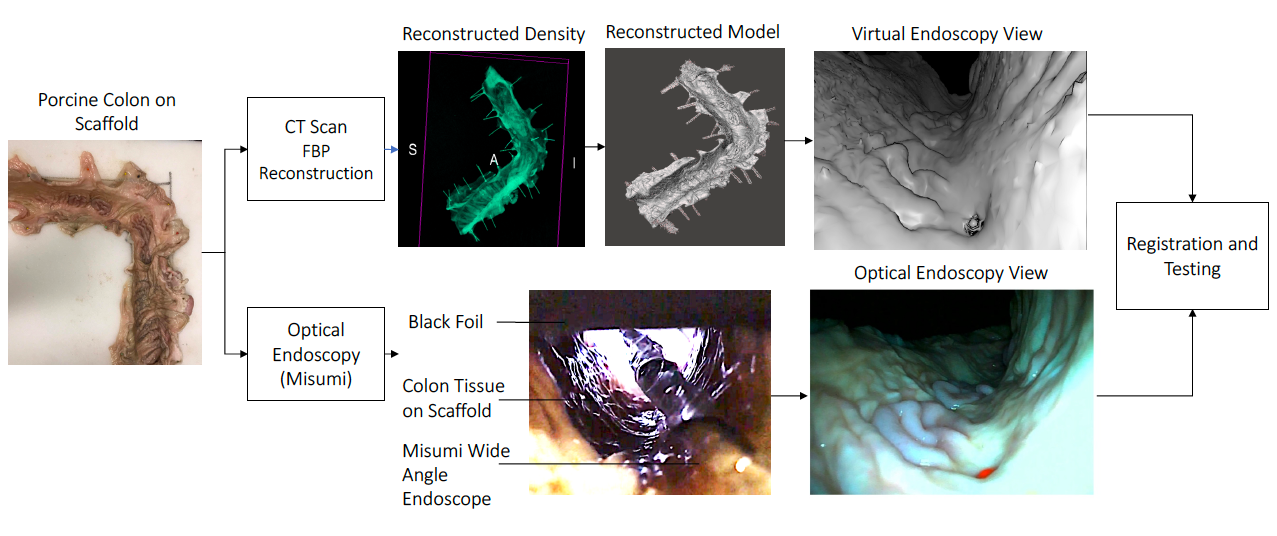}
\caption{Process of data generation from a porcine colon using reconstructed CT and optical endoscopy images. A porcine colon was dissected and placed on a scaffold. The scaffold was imaged on a cone-beam bench-top CT scanner and the 720 projections obtained were reconstructed using filtered back-projection. The CT reconstructed model was imaged with a virtual endoscope. The porcine colon was also imaged using an optical endoscope. The optical and virtual endoscopy views were registered and ground true depth for each optical endoscopy view was obtained.}
%\vspace*{2mm}
\label{fig_sim}
\end{figure*}

\begin{figure}
\centering
\includegraphics[width=7.0cm]{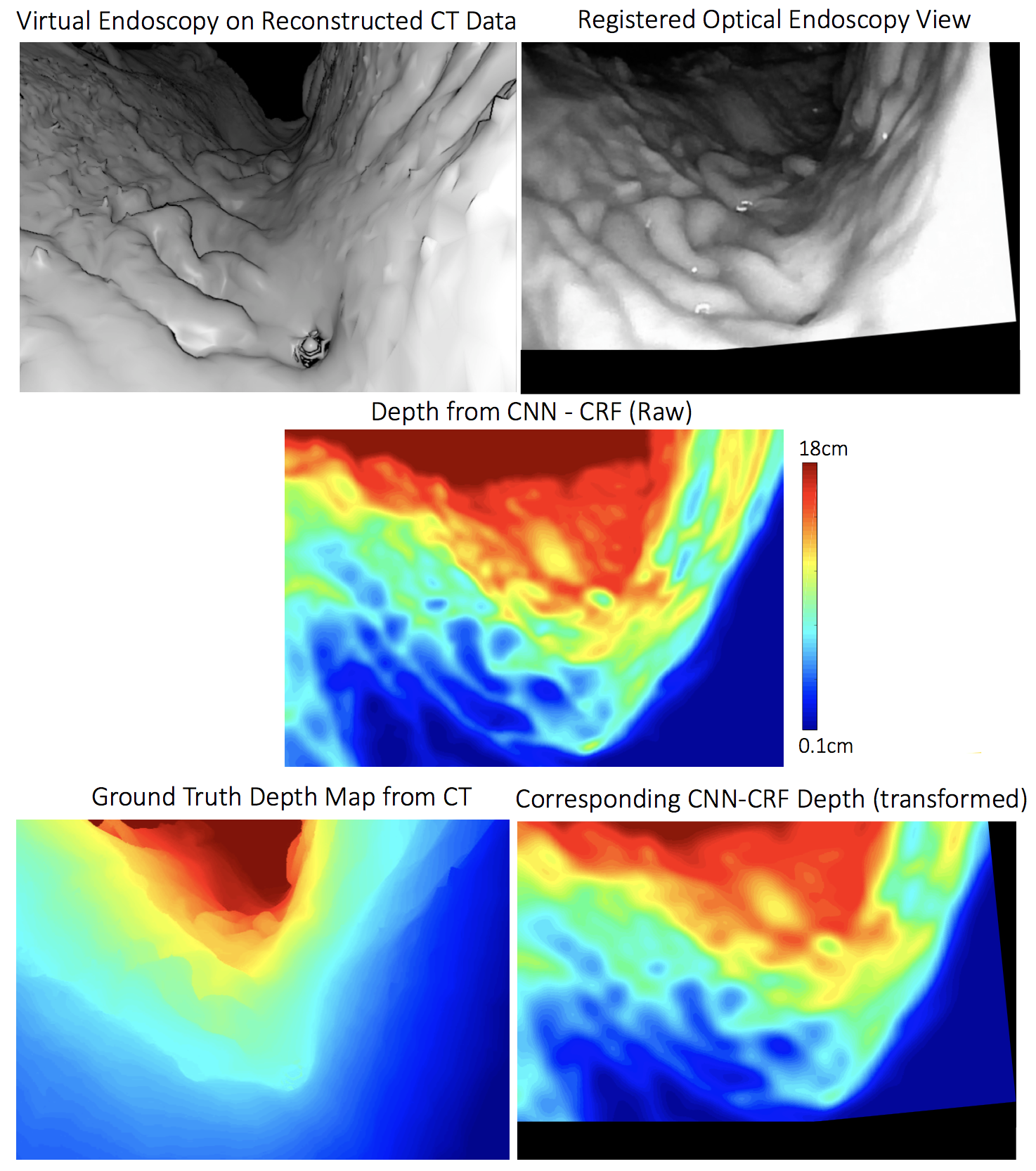}
\caption{Views from an optical and CT-virtual endoscopy on a porcine colon were registered using a one-plus-one evolutionary optimizer to generate ground truth depth for optical endoscopy images. The depth from CT-virtual endoscopy was calculated by measuring the distance from the virtual endoscope to each point appearing on the image. The depth from optical endoscopy was estimated using our network and was filtered like the reconstructed CT density for a fair comparison.}
\vspace*{-3.5mm}
\end{figure}

\subsection{Adversarial Training for Domain Adaptation}
\vspace{-0.8mm}

%Since our network is trained on images where high-level texture details are removed, this was done to suppress the network from training from texture and promote the inverse of intensity cue instead. We need a way to map the input images from a new endoscopic scene to a reduced texture representation as was the case with our training data. This was achieved by registering a dataset of 1460 images with optical endoscopic data collected on an anatomically accurate silicone phantom. The network used was a fully convolutional network presented for texture remapping in \cite{visentini2017deep}. Further details of this network are beyond the scope of this paper and have been mentioned in the supplementary material. 

Since our network was trained on synthetic data,  where low-level patient specific texture details are absent, we include a domain adaption step to test the network on real images. For new input images that contain this texture, we developed a network that transforms them to a synthetic like representation. This bridges the gap between real and synthetic domains. We use adversarial training between a discriminator network and a transformer network. This setup is based on recent advances in generative adversarial networks and adversarial training \cite{goodfellow2014generative,shrivastava2016learning}. The transformer network takes batches of synthetic images for unsupervised training and learns to remove patient-specific texture from the input images. The discriminator, which is embedded in the transformer's loss function, classifies the output as real or synthetic. Once the training reaches Nash equilibrium, the transformer is able to fool the discriminator every time and can perfectly transform a real image to its synthetic counterpart. To prevent the synthetic-like representation of a real image from deviating significantly from the original image we use a self-regularization term to preserve patient independent features. If $\theta_t$ and $\theta_d$ represent the learning parameters for the transformer and discriminator respectively, $\D_{\theta_d}$ represents the trained discriminator and $\T_{\theta_{t}}(\vec{x})$ represent the output of the transformer then the overall transformer loss term can be defined as,
\vspace{-2mm}
\begin{equation}\label{eq:Ttans_loss2}
  \begin{aligned}
 \Lno_{\T}(\theta_t) = -\sum_{i} \log(1-\D_{\theta_{d}}(\T_{\theta_{t}}(\vec{x}))) \\ + \lambda_3 \norm{\Phi(\T_{\theta_{t}}(\vec{x}))-\Phi(\vec{x})}_1.
  \end{aligned}
\end{equation}

The second term defines the $\ell_1$ self-regularization between the real endoscopy image and its synthetic-like counterpart. The $\Phi$ terms represent the feature transforms, for the sake of simplicity this was choosen to be a per-pixel loss. More details about this domain adaptation step and its implementation are beyond the scope of this paper and can be found here \cite{fm2017unsup}.

%that removes high spatial-frequency detail before applying the depth estimation network. This was achieved by training a new network on dataset of $1,460$ images with optical endoscopic data registered to a CT scan, collected on an anatomically realistic silicone phantom. We used a simple fully convolutional network previously developed for texture remapping in \cite{visentini2017deep}. Further details of this network are described in the supplementary material. 

\section{Experiments}

\subsection{Experimental Setup}

%We implemented the training networks using VLFeat MatConvNet \cite{vedaldi2015matconvnet} using MATLAB 2015a and CUDA 7.0. Training was done using K60 GPUs on the Maryland Advanced Computing Cluster (MARCC). Momentum was set at 0.9 and both weight decay parameters were set to 0.0005. The learning rate was initialized at 0.0001 and decrease by 40\% every 20 epoches as suggested in \cite{eigen_depth_2014}. A total of 300 epochs were ran and the epochs with least $log_{10}$ error were selected to avoid the selection of an over-fitted model. The endoscopy data was oversegmented into superpixels and corresponding ground true depth was assigned to each superpixel. The data was randomized to prevent the network from learning too many similar features quickly. The training takes about 82 hours and and depth for a single input frame can be predicted within about 1.4 sec.

We implemented the training networks using VLFeat Mat-ConvNet [48] using MATLAB 2017a and CUDA 8.0. The training data was prepared by oversegmenting each virtual endoscopy image into superpixels using SLIC \cite{ehlers_slic_1976} and corresponding ground truth depth was assigned to each superpixel. The data was randomized to prevent the network from learning too many similar features quickly. The network was pre-trained on synthetic colon data and fine tuned on colon phantom data. $55\%$ of the data was used for training and $40\%$ for validation and $5\%$ for testing. Training was done using K80 GPUs on the Maryland Advanced Computing Cluster (MARCC). Momentum was set at 0.9 as suggested in \cite{liu_learning_2016} and both weight decay parameters $(\lambda_1,\lambda_2)$ were set to 0.0007. The learning rate was initialized at 0.00001 and decreased by 20\% every 20 epoches. These parameters were tuned to achieve best results. A total of 300 epochs were run and the epochs with least $\log10$ error were selected to avoid the selection of an over-fitted model. 

%The training takes about 82 hours and depth for a single input frame can be predicted within about 1.4 sec <using x computer specs?>.

\subsection{Quantitative Evaluation Metrics}

%We evaluate the three datasets mentioned above in section II based on the following metrics, these metrics are inspired by other monocular depth estimation prior works mostly for mainstream vision. These metrics have been given below:

We evaluated the three datasets mentioned in section II	based on metrics used by other monocular depth estimation work \cite{ranftl2016dense,saxena2006learning,liu_learning_2016,li2015depth}, mostly for conventional vision. These metrics are:

\begin{enumerate}
\item Relative Error: $\frac{1}{N} \sum_{y} \frac{\lvert y_{gt}-y_{est} \rvert}{y_{gt}}$
\item Root Mean Square Error:$\sqrt{\frac{1}{N} \sum_{y} (y_{gt}-y_{est})^2}$
\item Average $log10$ Error: $\frac{1}{N} \sum_{y} \lvert \log_{10}y_{gt}-\log_{10}y_{est} \rvert$
%\item Accuracy with threshold:
\end{enumerate}

\begin{figure*}
\centering
\includegraphics[width=14cm]{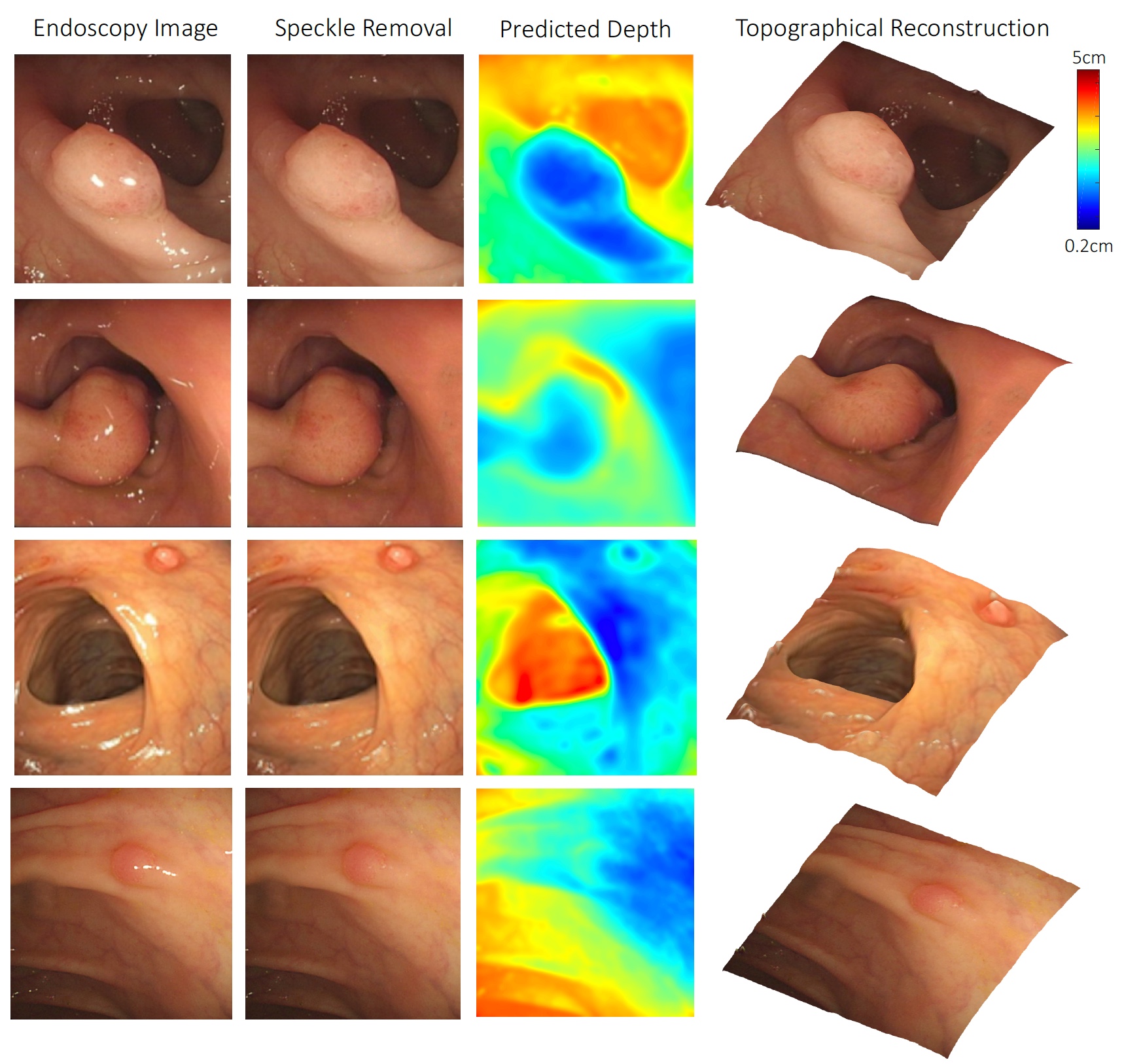}
\caption{Qualitative results showing predicted depth and topographical reconstructions for monocular endoscopy images.}
\label{fig_sim}
\end{figure*}

\begin{table}
\setlength{\tabcolsep}{12pt}

  \centering
  \begin{threeparttable}[b]
  \captionsetup{justification=centering,   textfont={sc}}
  \caption{Performance Evaluation for Synthetic Blender Generated Endoscopy Data}
  %\captionof{table}{Comparison of Regular and Graph Denoised (GD) Reconstructions}
  \label{tab:test2}
  \begin{tabular}{lllll}
  \toprule
    \multicolumn{1}{c}{Method} & {rel} & {$\log10$} & {rms}   \\
  %\multicolumn{1}{c}{} & {} & {} & {}  \\
   \cmidrule(l){1-4}
   Unary (FCN Only)       & 0.211       &0.094      &0.847          \\
   Unary (Smooth)         & 0.196       &0.083       &0.781          \\
   CNN-CRF	              & 0.152       &0.061       &0.612            \\
   \hline
  \end{tabular}
  %\begin{tablenotes}
  %  \item[*] Proposed Method
  %\end{tablenotes}
 \end{threeparttable}
\end{table}
\begin{table}
\setlength{\tabcolsep}{12pt}

  \centering
  \begin{threeparttable}[b]
  \captionsetup{justification=centering,   textfont={sc}}
  \caption{Performance Evaluation for Colon Phantom CT Rendered Virtual Endoscopy Data}
  %\captionof{table}{Comparison of Regular and Graph Denoised (GD) Reconstructions}
  \label{tab:test2}
  \begin{tabular}{llllll}
  \toprule
    \multicolumn{1}{c}{Method} & {rel} & {$\log10$} & {rms}  \\
  %\multicolumn{1}{c}{} & {} & {} & {}  \\

   \cmidrule(l){1-4}
   Unary (FCN Only)       & 0.227       &0.097       &0.907                \\
   Unary (Smooth)         & 0.216       &0.094       &0.884                 \\
   CNN-CRF	              & 0.183       &0.080       &0.753               \\
   \hline
  \end{tabular}
  %\begin{tablenotes}
  %  \item[*] Proposed Method
  %\end{tablenotes}
 \end{threeparttable}
\end{table}

\begin{table}
\setlength{\tabcolsep}{12pt}

  \centering
  \begin{threeparttable}[b]
  \captionsetup{justification=centering,   textfont={sc}}
  \caption{Performance Evaluation on Real Endoscopic Data Collected on a Porcine Colon Registered with CT}
  %\captionof{table}{Comparison of Regular and Graph Denoised (GD) Reconstructions}
  \label{tab:test2}
  \begin{tabular}{llllll}
  \toprule
    \multicolumn{1}{c}{Method} & {rel} & {$\log10$} & {rms}  \\
  %\multicolumn{1}{c}{} & {} & {} & {}  \\

   \cmidrule(l){1-4}
   Unary (FCN Only)       & 0.293       & 0.136      & 1.216            \\
   Unary (Smooth)         & 0.279       & 0.122      & 1.043            \\
   CNN-CRF	              & 0.242       & 0.098      & 0.973         \\
   \hline
  \end{tabular}
  %\begin{tablenotes}
  %  \item[*] Proposed Method
  %\end{tablenotes}
 \end{threeparttable}
\end{table}

\subsection{Comparative Analysis}

%It is not possible to compare our results directly with existing endoscopy depth estimation work because of the diversity of datasets and evaluation methods used. However, we do make a comparison with a plain FCN regression model, i.e., without the use of CRFs. This allows us to judge the benefit of using a graphical model. The CRF loss layer in the network is replaced with least squares regression. It should be noted that this architecture is similar to \cite{visentini2017deep}. However, we do not claim this to be a direct comparison between the two architectures because the data used is drastically different and their steup does not require the use of super-pixels.

It is not possible to compare our results directly with existing endoscopy depth estimation work because of the diversity of datasets and evaluation methods used. However, we do make a comparison with an FCN regression model that does not employ CRFs. Simple FCNs have recently been used for a variety of CV tasks including work for endoscopy \cite{visentini2017deep}. This comparison allows us to judge the benefit of using a graphical model. The CRF loss layer in the network is replaced with least squares regression. However, we do not claim this to be a direct comparison with \cite{visentini2017deep} or other works because the data used is drastically different and their setup does not use of super-pixels.

\begin{figure}
\centering
\includegraphics[width=8.5cm]{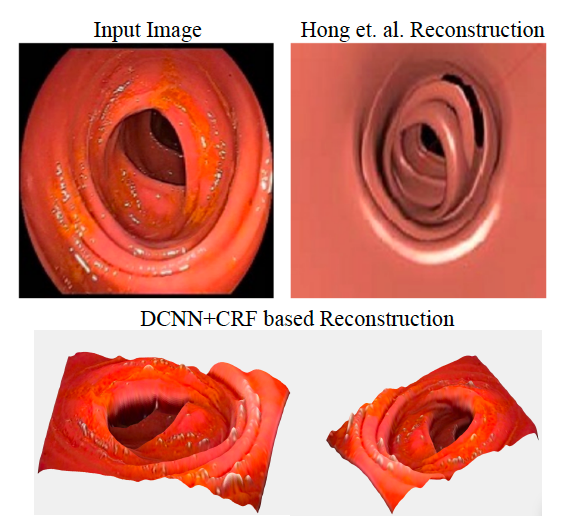}
\caption{Comparison of our proposed depth estimation and reconstruction method with a reconstruction from \cite{hong_3d_2014}. Input image and 3D reconstruction was taken from \cite{hong_3d_2014}.}
\label{fig_sim}
\vspace{-3.5mm}
\end{figure}
\vspace{-1.5mm}
\subsection{Results with Synthetic Colon and Phantom Virtual Endoscopy Data}

%The trained network was tested on a set of images that were part of the synthetic colon data and CT colon phantom data not used for training.  Using randomly selected 10,000 test images we verify the overall quality of our network for images which are very similar to the training data. Table I, II and Fig. 5 show these results in detail.

The trained network was tested on images from the synthetic colon and silicone colon phantom that were not used for training (Fig. 6). Using 10,000 randomly-selected test images we observe that the accuracy of the network improves by every metric with the CNN-CRF method for images which are very similar to the training data (Table II, III).
\vspace{-1.5mm}
\subsection{Results with Porcine Colon Real Endoscopy Data}

%As mentioned earlier, to validate the effectiveness of our trained models on real tissue we collected data on a a porcine colon and placed it on a scaffold with an anatomically accurate bend followed by a CT scan. The CT reconstructions from the scan were then registered with optical endoscopy images which could give us ground true test data. Fig. 6 shows the validation flow in detail and Table III presents the results. Besides the error from the network there are several sources of inaccuracy that contribute to the overall error. These sources include errors from significant manipulation of raw CT data during the reconstruction, refinement and filtering process. Moreover, the CT data also includes streaking artifacts due to non-uniform x-ray absorption specifically around the metallic pin fiducial markers added during data collection. More error sources include inconsistency of the stepper motor which rotated the scaffold and errors related to registering the virtual endoscopy CT view with the optical endoscopy image. 
As mentioned earlier optical and virtual endoscopy views from a CT reconstruction were registered to get ground truth depth maps for optical endoscopy images (Fig. 7). For a fair comparison, the depth map from endoscopy was filtered through the same pipeline of filters used for the CT reconstruction. Only registered regions of the two depth maps were compared. Representative images from this process are shown in Fig. 8 and a the algorithm performance on $1,460$ porcine colon images is summarized in Table IV.

%Besides the inherent accuracy of the network there are several sources of error. Significant manipulation of raw CT data during the reconstruction, refinement and filtering process. Moreover, the CT data also includes streaking artifacts due to non-uniform x-ray absorption specifically around the metallic pin fiducial markers added during data collection. More error sources include inconsistency of the stepper motor which rotated the scaffold and errors related to registering the virtual endoscopy CT view with the optical endoscopy image.

\subsection{Qualitative Results with Real Human Endoscopy Data}

%We tested the trained network on real data from endoscopy images available form NIH and the MICCAI endoscopy challenge databases. The results have been shown in Fig. 7, it can clearly be seen that the network can regress  course depth maps. These depth maps were then used to reconstruct the topography of the surface of the colon. Fig. 8 shows a comparison of our method with 3D monocular reconstructions from Hong et.al. \cite{hong_3d_2014}.

We tested the trained network on real data from colonoscopy images available from the NIH and the MICCAI endoscopy challenge databases \cite{tajbakhsh2016automated,silva2014toward,bernal2015wm}. The results in Fig. 9, show that the network can regress coarse depth maps that match intuitive cues from the image. These depth maps were then used to reconstruct the topography of the surface of the colon. Fig. 10 compares our method with 3D monocular reconstruction from the tubular assumption-based approach presented by Hong \textit{et al.} \cite{hong_3d_2014}.

\section{Conclusions}

%In this paper we propose an architecture for monocular endoscopy depth estimation and topographical reconstruction. We used a joint CNN and CRF-based framework to regress depth from monocular images. Unlike previous approaches this method does not require geometric assumptions. The network was trained using 200,000 images from synthetically generated data and reconstructed CT data imaged using a virtual endoscope. To the best of our knowledge this is the first work which trains a large scale medical imaging network from synthetically generated/rendered data that would otherwise be quite tedious to obtain due to limitations of endoscope hardware and size. We validate our network by generating a test dataset using a real porcine colon and mounting it on a scaffold followed by CT and optical endoscopy data collection from the scaffold. Monocular endoscopic depth and topographical structure of the surface of the colon fits the broader framework of 3D endoscopy research and have the potential to significantly improve CAD algorithms for detection, segmentation and classifications of lesions. 

This paper presents a novel architecture for monocular endoscopy depth estimation and topographical reconstruction that uses the advantages of a joint CNN and CRF-based framework. Unlike previous, approaches this method does not require geometric assumptions. The network was trained using 200,000 images from synthetically-generated data and CT-reconstructions imaged using a virtual endoscope. To the best of our knowledge, this is the first work which trains a network from a large set of synthetically generated and rendered medical images. This is a particularly relevant approach to 3D endoscopy applications because, despite the clinical need, there are no practical alternatives to acquiring large datasets of real endoscopy images with corresponding ground truth. We validate our network on real colon tissue and endoscopy by generating a test dataset using a porcine colon and mounting it on a scaffold followed by CT and registered optical endoscopy. Our work adds to the active area of 3D endoscopy research and has the potential to improve CAD algorithms for detection, segmentation and classifications of lesions. 

The limitations of the current method include artifacts due to specular reflections, cases where inverse of intensity might not be the major cue and instances where the pairwise similarities can give rise to artifacts. Moreover, there were several sources of testing errors beyond the inherent accuracy of the network. The reconstruction, refinement, and filtering of the raw CT data all contribute to inaccuracies in the depth map used for ground truth. The CT data also includes streaking artifacts due to non-uniform x-ray absorption, specifically around the metallic pin fiducial markers. More error sources include inconsistency of the stepper motor which rotated the scaffold and errors related to registering the virtual endoscopy CT view with the optical endoscopy image.

Our future work will focus on generalizing the concept of synthetic data generation for medical images and utilizing depth estimation as an additional cue for other endoscopy applications. The proposed network can also be used as an initialization for future deep learning-oriented endoscopy applications.

\vspace{-4mm}

\section{Acknowledgments}

The authors thank Dr. J. Webster Stayman and Mr. Steven Tilley II (Department of Biomedical Engineering, Johns Hopkins University, Baltimore, MD) for collecting cone-beam CT data on the porcine colon, and Dr. Jeffrey Siewerdsen for his insightful feedback on the manuscript. The authors  also thank the staff at Maryland Advanced Computing Cluster (MARCC) for their efficient technical support and training.

% that's all folk
\vspace{-4mm}

% Can use something like this to put references on a page
% by themselves when using endfloat and the captionsoff option.
%\FloatBarrier
\ifCLASSOPTIONcaptionsoff
 % \newpage
\fi

%\balance

\bibliographystyle{IEEEtran}
% argument is your BibTeX string definitions and bibliography database(s)
\bibliography{bare_jrnl_mm1.bbl}

\end{document}